\documentclass[10pt, a4paper]{article}

\usepackage{lrec-coling2024} 

\usepackage{hyperref}
\usepackage{url}
\usepackage{bbm}
\usepackage{bm}
\usepackage{booktabs}
\usepackage{amsmath}
\usepackage{amssymb}
\usepackage{tabu}
\usepackage{CJKutf8}

\title{Release of Pre-Trained Models for the Japanese Language}

\name{Kei Sawada, Tianyu Zhao, Makoto Shing, Kentaro Mitsui, \\
\large{\textbf{Akio Kaga, Yukiya Hono, Toshiaki Wakatsuki, Koh Mitsuda}}}

\address{rinna Co., Ltd., Tokyo, Japan \\
          keisawada@rinna.co.jp\\}

\abstract{ 
AI democratization aims to create a world in which the average person can utilize AI techniques. 
To achieve this goal, numerous research institutes have attempted to make their results accessible to the public. 
In particular, large pre-trained models trained on large-scale data have shown unprecedented potential, and their release has had a significant impact. 
However, most of the released models specialize in the English language, and thus, AI democratization in non-English-speaking communities is lagging significantly. 
To reduce this gap in AI access, we released Generative Pre-trained Transformer (GPT), Contrastive Language and Image Pre-training (CLIP), Stable Diffusion, and Hidden-unit Bidirectional Encoder Representations from Transformers (HuBERT) pre-trained in Japanese. 
By providing these models, users can freely interface with AI that aligns with Japanese cultural values and ensures the identity of Japanese culture, thus enhancing the democratization of AI. 
Additionally, experiments showed that pre-trained models specialized for Japanese can efficiently achieve high performance in Japanese tasks.

\\ \newline 
}

\begin{document}

\maketitleabstract

\section{Introduction}
\label{sec:intro}
As AI technology advances, the idea of ``AI democratization,'' which aims to create a world where everyone can easily use AI, has become widely popular. 
To contribute to AI democratization, many research institutions and companies are publicly releasing their latest methods, source codes, databases, and pre-trained models. 
Such steps are essential for supporting the rapid development of AI technology in the future. 

Recently, methods using large-scale pre-trained models based on massive training data have achieved significant results and have become mainstream.
The advent of self-supervised learning, which generates pseudo-ground-truth labels from training data, coupled with the introduction of the Transformer architecture~\citep{DBLP:conf/nips/VaswaniSPUJGKP17}, which enables efficient and accurate model training from massive data, has made large-scale modeling possible. 
The Generative Pre-trained Transformer (GPT,~\citealt{radford2018improving}) series engendered a breakthrough in text generation using self-supervised learning and Transformer architectures by discovering a scaling law suggesting that the performance improves as the model size, amount of training data, and computation increase~\citep{kaplan2020scaling}.
As a result, the size of the pre-trained models has escalated dramatically in the text domain as well as the image and speech domains.
However, training high-performance pre-trained models incurs significant costs, such as creating training corpora and securing computational resources, making it infeasible for everyone to undertake pre-training easily. 
Fortunately, there is an active trend of releasing pre-trained models on platforms such as Hugging Face, and such models are now available. 

While there is vibrant activity in the publishing of pre-trained models, many pre-trained models targeting languages are specialized for English. 
Consequently, AI democratization lags in non-English-speaking regions compared with English-speaking regions. 
Research is underway on multilingual models that support several languages. 
However, these multilingual models tend to have an increased number of parameters and often underperform compared with models specialized for a particular language given a fixed compute budget~\citep{lin2022few}.

To address this issue in Japanese, we released pre-trained models optimized for Japanese on Hugging Face. 
By providing pre-trained models specialized for Japanese, we hope that users can freely access a model that aligns with Japanese cultural values but also ensures the identity of Japanese culture, leading to a more inclusive AI democratization that does not solely lean towards English-centric perspectives.

\vspace{-2mm}
\section{Japanese Pre-Trained Models}
\vspace{-1mm}
\begin{table*}[!t]
\centering
\begin{tabular}{lrrr}
\toprule
\textbf{Pre-trained model} & \textbf{Model size} & \textbf{License} & \textbf{Release date}\\
\midrule
Language model & & & \\
\quad {\it rinna/japanese-gpt2-xsmall} & 37M & MIT & August 2021\\
\quad {\it rinna/japanese-gpt2-small} & 110M & MIT & August 2021\\
\quad {\it rinna/japanese-gpt2-medium} & 336M & MIT & April 2021\\
\quad {\it rinna/japanese-gpt-1b} & 1.3B & MIT & January 2022\\
\quad {\it rinna/japanese-gpt-neox-small} & 110M & MIT & September 2022\\
\quad {\it rinna/japanese-gpt-neox-3.6b} & 3.6B & MIT & May 2023\\
\quad {\it rinna/bilingual-gpt-neox-4b} & 4B & MIT & July 2023\\
\midrule
Language-image model & & & \\
\quad {\it rinna/japanese-clip-vit-b-16} & 197M & Apache 2.0 & May 2022\\
\quad {\it rinna/japanese-cloob-vit-b-16} & 197M & Apache 2.0 & May 2022\\
\quad {\it rinna/japanese-stable-diffusion} & 1.1B & {CreativeML OpenRAIL M} & September 2022\\
\midrule
Speech model & & & \\
\quad {\it rinna/japanese-hubert-base} & 95M & Apache 2.0 & April 2023\\
\bottomrule
\end{tabular}
\caption{Released pre-trained models in the Japanese language.}
\label{table:models}
\end{table*}

We built pre-trained models appropriate for the Japanese language and culture and released them in Hugging Face\footnote{\url{https://huggingface.co/rinna}}. 
Table~\ref{table:models} presents an overview of the released pre-trained models we have released by September 2023.
These models have fewer restrictive licenses, thereby allowing their wide use. 
In fact, between April 2021 and September 2023, these models were downloaded over four million times from Hugging Face. 
The details and specifics of these models are discussed in Sections~\ref{sec:lm} to~\ref{sec:sm}.

\section{Language Models}
\label{sec:lm}
\subsection{GPT}
\subsubsection{Overview}
The Generative Pre-trained Transformer (GPT,~\citealt{radford2018improving}) is an autoregressive language model composed of an input embedding layer, stacked Transformer layers~\citep{DBLP:conf/nips/VaswaniSPUJGKP17}, and an output classification layer. It models $p(\mathbf{x})$, the probability of a sequence of text tokens $\mathbf{x}=[x_1, \cdots, x_{|\mathbf{x}|}]$, as factorized token-level probabilities, and then pre-trains a GPT model to minimize the negative log-likelihood (NLL) $\mathcal{L}_{\text{NLL}}$.
\begin{align}
    p(\mathbf{x}) &= p(x_1) p(x_2 | x_1) \cdots p(x_{|\mathbf{x}|} | x_{:|\mathbf{x}|-1}), \\
    \mathcal{L}_{\text{NLL}} &= - \log p(\mathbf{x}).
\end{align}

GPT-NeoX~\citep{DBLP:journals/corr/abs-2204-06745} is a GPT variant that uses a modified architecture for the Transformer layer and an alternative position encoding mechanism called rotary embedding~\citep{DBLP:journals/corr/abs-2104-09864} as a substitute for original learnable position embeddings.

For the most capable models, we also released their instruction-following versions, which were trained using either Supervised Fine-Tuning (SFT) or Reinforcement Learning from Human Feedback (RLHF,~\citealt{DBLP:conf/nips/Ouyang0JAWMZASR22}) via the Proximal Policy Optimization (PPO,~\citealt{DBLP:journals/corr/SchulmanWDRK17}) algorithm in addition to SFT.

\subsubsection{Training Data}
For Japanese-specific GPT models, we used Wikipedia, the CC-100~\citeplanguageresource{DBLP:conf/acl/ConneauKGCWGGOZ20}, and the mC4~\citeplanguageresource{DBLP:journals/jmlr/RaffelSRLNMZLL20} datasets for pre-training. For bilingual English-Japanese GPT models, we additionally used the Pile~\citeplanguageresource{DBLP:journals/corr/abs-2101-00027} and Redpajama~\citeplanguageresource{together2023redpajama} datasets. The instruction-following models were trained on the Japanese translation of the Anthropic HH~\citeplanguageresource{DBLP:journals/corr/abs-2204-05862}, the SHP~\citeplanguageresource{pmlr-v162-ethayarajh22a}, and the FLAN~\citeplanguageresource{DBLP:conf/iclr/WeiBZGYLDDL22} datasets.

Tokenizers of the GPT models are trained via SentencePiece~\citep{sp_tokenizer}. 
Their vocabulary sizes vary from 32000 to 65536. While the tokenizers for Japanese-only models are trained from Japanese corpora, the tokenizer of \textit{bilingual-gpt-neox-4b} is trained from a mixture of Japanese and English corpora for a better coverage of English tokens. 

\subsubsection{Experiments}
We conducted few-shot evaluations of the GPT models to assess their performance on Japanese tasks. 
We used the JP Language Model Evaluation Harness\footnote{\url{https://github.com/Stability-AI/lm-evaluation-harness/tree/jp-stable}} benchmark for evaluation. 
We conducted a comparison with {\it meta/llama-7b}~\citep{touvron2023llama}, {\it meta/llama2-7b}, and {\it meta/llama2-7b-chat}~\citep{touvron2023llama2}, which were primarily trained using English data.

Table~\ref{table:lm-eval-harness} lists the average scores for the jcommonsenseqa, jnli, marc-ja, jsquad~\citeplanguageresource{jglue}, xwinograd~\citeplanguageresource{muennighoff-etal-2023-crosslingual}, jaqket-v2\footnote{\url{https://sites.google.com/view/project-aio/competition2}}, xlsum-ja~\citeplanguageresource{xlsum}, and mgsm~\citeplanguageresource{shi2023language} tasks. 
The few-shot numbers were 3, 3, 3, 2, 0, 1, 1, and 5. 
Our {\it rinna/japanese-gpt-neox-3.6b} and {\it rinna/bilingual-gpt-neox-4b} pre-trained models outperformed {\it meta/llama-7b}, and instruction tuning via SFT or PPO significantly improved their capability. 
By specializing in Japanese, good performance was achieved while keeping the number of parameters low.
We refer the readers to rinna's language model benchmark\footnote{\url{https://rinnakk.github.io/research/benchmarks/lm/index.html} Due to the update of the evaluation code base, the latest benchmark adopts a different evaluation setting than that used in this paper. 
The results presented in this paper can be found in the benchmark spreadsheet on the \textit{20231031} tab.} for detailed benchmark results. 

\begin{table}[!t]
\centering
\begin{tabular}{lr}
\toprule
\textbf{Pre-trained model} & \hspace{-10mm} \textbf{Average score} \\
\midrule
{\it rinna/japanese-gpt2-xsmall} & 26.63 \\
{\it rinna/japanese-gpt2-small} & 27.33 \\
{\it rinna/japanese-gpt2-medium} & 28.33 \\
{\it rinna/japanese-gpt-1b} & 32.21 \\
{\it rinna/japanese-gpt-neox-small} & 30.11 \\
{\it rinna/japanese-gpt-neox-3.6b} & 36.60 \\
{\it rinna/bilingual-gpt-neox-4b} & 38.29 \\
\midrule
{\it meta/llama-7b} & 33.28 \\
{\it meta/llama2-7b} & 42.97 \\
{\it meta/llama2-7b-chat} & 41.31 \\
\midrule
{\it rinna/japanese-gpt-neox-3.6b-sft} & 45.24 \\
{\it rinna/japanese-gpt-neox-3.6b-ppo} & 46.37 \\
{\it rinna/bilingual-gpt-neox-4b-sft} & 47.65 \\
{\it rinna/bilingual-gpt-neox-4b-ppo} & 47.33 \\
\bottomrule
\end{tabular}
\caption{Language model evaluation on the JP Language Model Evaluation Harness.}
\label{table:lm-eval-harness}
\end{table}

\section{Language-Image Models}
\label{sec:lim}
\subsection{CLIP} \label{section:clip}

\subsubsection{Overview}
Contrastive Language-Image Pre-training (CLIP, \citealt{clip}) connects visual concepts with natural language in the embedding space.
It comprises a pair of text and image encoders and is trained by minimizing contrastive loss.
The Contrastive Leave One Out Boost (CLOOB, \citealt{cloob}) demonstrated a better zero-shot performance than the original CLIP by introducing a novel loss function termed InfoLOOB. 

To train the Japanese-specific CLIP efficiently, we applied Locked-image Tuning (LiT, \citealt{lit}), in which both encoders were initialized using separate pre-trained models, and only the text encoder was trained.
We used the pre-trained 12-layer 16$\times$16-patch-size AugReg Vision Transformer~\citep{vit,augreg} for the image encoder, and randomly initialized 12-layer Bidirectional Encoder Representations from Transformers (BERT,~\citealt{DBLP:conf/naacl/DevlinCLT19}) with a SentencePiece tokenizer~\citep{sp_tokenizer} for the text encoder. 

\subsubsection{Training Data}
Owing to the absence of a large-scale dataset with Japanese captions, we used CC12M~\citeplanguageresource{cc12m}.
We translated all the English captions into Japanese. 
For data augmentation, we generated captions using Bootstrapping Language-Image Pre-training (BLIP, \citealt{blip}) trained on an English dataset.

\begin{table}[!t]
    \centering
    \begin{tabular}{lr}
         \toprule
         \textbf{Pre-trained model}  & \textbf{Accuracy} \\
         \midrule
         {\it laion/clip-base} & 38.00 \\
         {\it laion/clip-large} & 53.09 \\
         \midrule
         {\it rinna/japanese-clip-vit-b-16} & 50.69  \\
         {\it rinna/japanese-cloob-vit-b-16} & 54.64 \\
         \bottomrule
    \end{tabular}
    \caption{ImageNet image classification accuracy in a zero-shot setting.}
    \label{table:imagenet}
\end{table}

\subsubsection{Experiments}
We evaluated CLIP for ImageNet~\citep{imagenet} zero-shot image classification.
We used open-sourced Japanese class names\footnote{\url{https://gist.github.com/PonDad/4dcb4b242b9358e524b4ddecbee385e9}}.
Additionally, we created 37 Japanese templates from 80 English templates by deduplicating captions that had the same meaning in Japanese.
We compared our models with open-source multilingual CLIP models~\citep{openclip} trained on full LAION-5B~\citeplanguageresource{laion5b}.

Table~\ref{table:imagenet} shows the top-1 accuracy for each model. 
Our {\it rinna/japanese-cloob-vit-b-16} performed the best and achieved state-of-the-art accuracy. 
This is because, even with a limited amount of training data, the model can be efficiently trained by specializing in a specific language. 

\subsection{Stable Diffusion}

\subsubsection{Overview}
Stable Diffusion (SD) facilitates high-quality image generation using simple text prompts.
It is based on the Latent Diffusion Model (LDM, \citealt{ldm}), which comprises three main components: a Variational AutoEncoder (VAE, \citealt{vae}), a text encoder, and U-Net~\citep{10.1007/978-3-319-24574-4_28}.

To train the Japanese-specific SD (JSD), we fine-tuned {\it CompVis/stable-diffusion-v1-4}\footnote{\url{https://huggingface.co/CompVis/stable-diffusion-v1-4}} trained on the English dataset.
We applied two training stages following the concept of Pretraining-based Image-To-Image translation (PITI, \citealt{piti});
The text encoder was trained solely with U-Net fixed in the first stage and jointly trained in the second stage.

\subsubsection{Training Data}
We used approximately 100 million images with Japanese captions, including the Japanese subset LAION-5B~\citeplanguageresource{laion5b}. 
To ensure data quality, we employed our {\it rinna/japanese-cloob-vit-b-16} introduced in Section~\ref{section:clip} to calculate the similarity scores between images and their captions, and samples with scores below a certain threshold were removed.

\subsubsection{Experiments}
\begin{figure}[!t]
    \centering
    \begin{tabular}{cc}
        \hspace{-0.5cm}{\scriptsize SD}\hspace{-0.6cm}
        &
        \begin{minipage}{7cm}
      \centering
      \includegraphics[width=\linewidth]{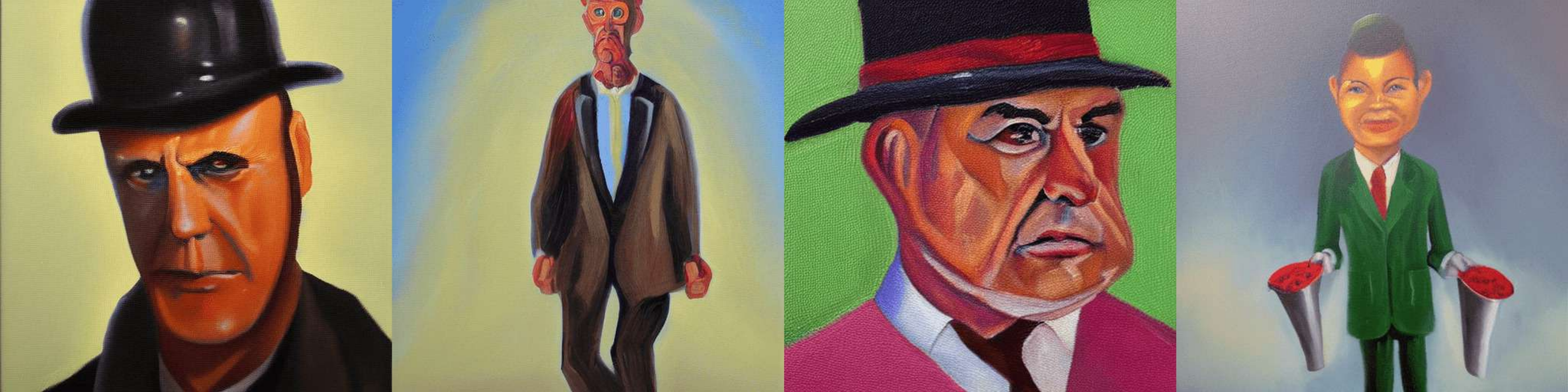}
        \end{minipage} \\
        \hspace{-0.5cm}
        \begin{tabular}{c}
        {\scriptsize JSD} \\
        {\scriptsize (1st)}
        \end{tabular}
        \hspace{-0.6cm}
        &
        \begin{minipage}{7cm}
          \centering
          \includegraphics[width=\linewidth]{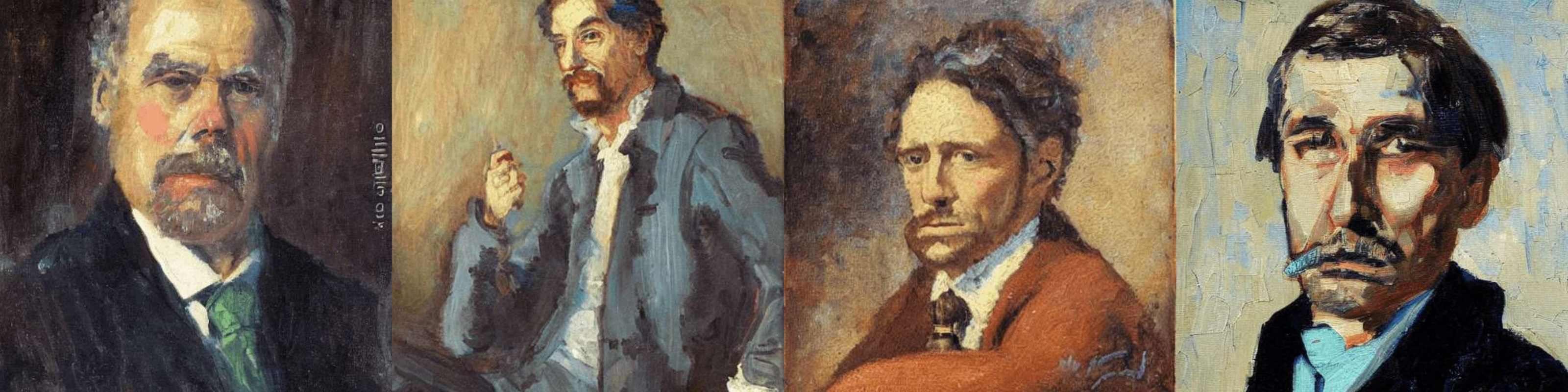}
        \end{minipage}  \\
        \hspace{-0.5cm}
        \begin{tabular}{c}
        {\scriptsize JSD} \\
        {\scriptsize (2nd)}
        \end{tabular}
        \hspace{-0.6cm}
        & 
        \begin{minipage}{7cm}
          \centering
          \includegraphics[width=\linewidth]{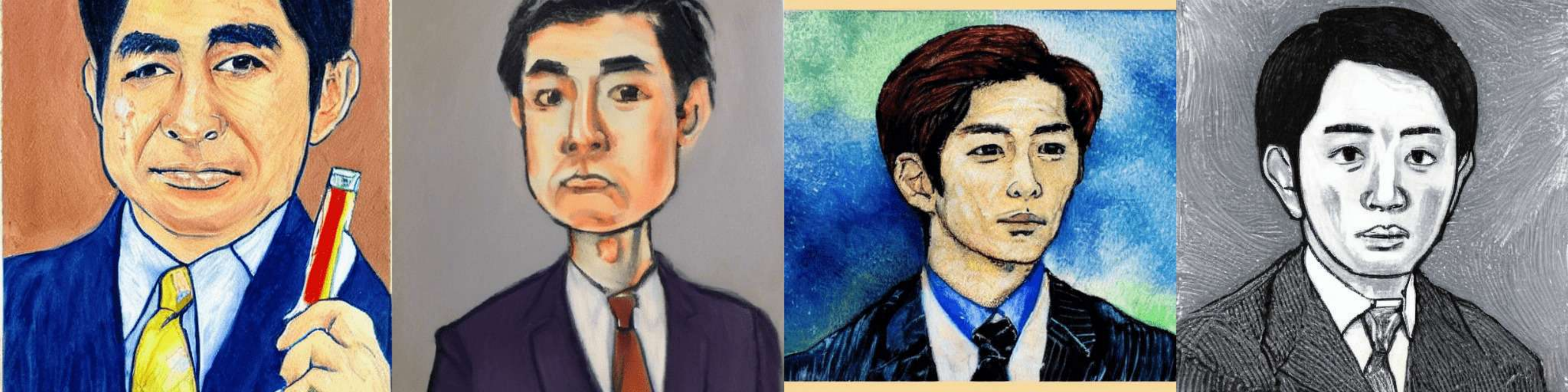}
    \end{minipage} 
    \end{tabular}
    
    \caption{Outputs for the text prompt ``salary man, oil painting''. For JSD, the translation in Japanese ``\begin{CJK}{UTF8}{ipxm}サラリーマン 油絵\end{CJK}'' was used. \label{fig:sd-salaryman}}
\end{figure}

We used Japanglish ``salary man'', which is commonly visualized as a man in a suit, as the text prompt for the evaluation. 
Figure~\ref{fig:sd-salaryman} shows the results. 
The original SD failed to accurately interpret such distinctive Japanese terms. 
In the first stage, JSD understood the prompt's meaning, but the generated images depicted businessmen with Western features because U-Net had not been updated. 
JSD at the second stage ({\it rinna/japanese-stable-diffusion}), JSD successfully generated images of businessmen with Japanese features. 
Using images reflecting Japanese culture as the training data, we were able to construct a model consistent with Japanese cultural identity.

\section{Speech Models}
\label{sec:sm}
\subsection{HuBERT}
\subsubsection{Overview}
The Hidden-unit BERT (HuBERT,~\citealt{hsu2021hubert}) is a pre-trained model that can learn self-supervised speech representations.
HuBERT comprises two main components: a convolutional waveform encoder and a BERT encoder~\citep{DBLP:conf/naacl/DevlinCLT19}.
HuBERT is trained with a BERT-like masked prediction objective: a portion of the encoded speech feature sequence is randomly masked, and a label corresponding to the masked portion is predicted from the unmasked portion.
However, because speech signals, unlike text, are continuous-valued sequences, the model is trained by targeting discrete pseudo-labels obtained from speech using offline $k$-means clustering.

\subsubsection{Training Data}
We used the ReazonSpeech corpus \citeplanguageresource{yin20222reazonspeech}, a 19,000-hour speech corpus collected from Japanese TV programs with 16~kHz sampling.
To generate pseudo-labels, we ran $k$-means clustering with 100 clusters on 39-dimensional Mel-frequency cepstral coefficient features for the first iteration of HuBERT training and 500 clusters on the latent features extracted from the 6-th Transformer layers' of the first iteration for the second iteration of HuBERT training.

\subsubsection{Experiments}
We evaluated the performance of the pre-trained HuBERT model for Japanese Automatic Speech Recognition (ASR).
We used Corpus of Spontaneous Japanese~\citeplanguageresource{maekawa2000spontaneous}.
Two training subset sizes were prepared: core data only (approximately 32 h) and all data (approximately 552 h).
ASR fine-tuning using the Connectionist Temporal Classification (CTC,~\citealt{graves2006connectionist}) loss was performed as described in~\citep{hsu2021hubert}.
The final projection layer was replaced with a softmax layer before ASR fine-tuning.
The target vocabulary included 40 Japanese phonemes, a word boundary symbol, and a special CTC blank symbol.
The public HuBERT model {\it meta/hubert-base-ls960}\footnote{\url{https://huggingface.co/facebook/hubert-base-ls960}}, pre-trained with 960 hours English speech from Librispeech~\citeplanguageresource{panayotov2015librispeech}, was used for comparison.
In this study, we used the beam search with a beam size of 20 without a language model.

\begin{table}[!t]
    \centering
    \begin{tabular}{lrrr}
        \toprule
        \textbf{Pre-trained model} & \textbf{Eval1} & \textbf{Eval2} & \textbf{Eval3} \\
        \midrule
        \multicolumn{4}{l}{{\it meta/hubert-base}} \\
        \quad 32-hour labeled & 13.12 & 10.33 & 10.66 \\
        \quad 552-hour labeled & 7.88 & 5.66 & 6.48 \\
        \midrule
        \multicolumn{4}{l}{{\it rinna/japanese-hubert-base}} \\
        \quad 32-hour labeled & 9.30 & 7.07 & 6.87 \\
        \quad 552-hour labeled & 5.72 & 4.45 & 4.73 \\
        \bottomrule
    \end{tabular}
    \caption{
        The word error rates for the fine-tuned HuBERT models with the different sizes of data.
    }
    \label{table:asr}
\end{table}

The results are presented in Table~\ref{table:asr}.
For both sizes of labeled data, the {\it rinna/japanese-hubert-base} outperformed the {\it meta/hubert-base-ls960}.
This result indicates that the pre-trained HuBERT model trained with a large Japanese speech corpus has the potential to provide better performance in Japanese speech-processing tasks.

\section{Conclusion}
\label{sec:conc}
Aiming to advance AI democratization, this paper discusses the released Japanese GPT, CLIP, Stable Diffusion, and HuBERT models.
Experiments with GPT, CLIP, and HuBERT showed that pre-trained models specialized for Japanese can efficiently achieve high performance in Japanese tasks. 
Additionally, the Stable Diffusion results indicate that it handles Japanese input and produces output that reflects Japanese culture. 
Pre-trained models are continuously refined, and technically challenging tasks for improving them have now become achievable. 
We plan to continue releasing pre-trained models to further contribute to technological progress.

\section{Bibliographical References}
\label{sec:reference}
\bibliographystyle{lrec-coling2024-natbib}
\bibliography{mybib}

\section{Language Resource References}
\bibliographystylelanguageresource{lrec-coling2024-natbib}
\bibliographylanguageresource{languageresource}

\clearpage
\appendix
\begin{table*}[!t]
\centering
\begin{tabular}{lrrr}
\toprule
\textbf{Pre-trained model} & \textbf{License} & \textbf{Release date}\\
\midrule
Language model & & & \\
\quad {\it \href{https://huggingface.co/rinna/japanese-roberta-base}{rinna/japanese-roberta-base}} & MIT & August 2021\\
\quad {\it \href{https://huggingface.co/rinna/japanese-gpt2-xsmall}{rinna/japanese-gpt2-xsmall}} & MIT & August 2021\\
\quad {\it \href{https://huggingface.co/rinna/japanese-gpt2-small}{rinna/japanese-gpt2-small}} & MIT & August 2021\\
\quad {\it \href{https://huggingface.co/rinna/japanese-gpt2-medium}{rinna/japanese-gpt2-medium}} & MIT & April 2021\\
\quad {\it \href{https://huggingface.co/rinna/japanese-gpt-1b}{rinna/japanese-gpt-1b}} & MIT & January 2022\\
\quad {\it \href{https://huggingface.co/rinna/japanese-gpt-neox-small}{rinna/japanese-gpt-neox-small}} & MIT & September 2022\\
\quad {\it \href{https://huggingface.co/rinna/japanese-gpt-neox-3.6b}{rinna/japanese-gpt-neox-3.6b}} & MIT & May 2023\\
\quad {\it \href{https://huggingface.co/rinna/japanese-gpt-neox-3.6b-instruction-sft}{rinna/japanese-gpt-neox-3.6b-instruction-sft}} & MIT & May 2023\\
\quad {\it \href{https://huggingface.co/rinna/japanese-gpt-neox-3.6b-instruction-sft-v2}{rinna/japanese-gpt-neox-3.6b-instruction-sft-v2}} & MIT & May 2023\\
\quad {\it \href{https://huggingface.co/rinna/japanese-gpt-neox-3.6b-instruction-ppo}{rinna/japanese-gpt-neox-3.6b-instruction-ppo}} & MIT & May 2023\\
\quad {\it \href{https://huggingface.co/rinna/bilingual-gpt-neox-4b}{rinna/bilingual-gpt-neox-4b}} & MIT & July 2023\\
\quad {\it \href{https://huggingface.co/rinna/bilingual-gpt-neox-4b-8k}{rinna/bilingual-gpt-neox-4b-8k}} & MIT & July 2023\\
\quad {\it \href{https://huggingface.co/rinna/bilingual-gpt-neox-4b-instruction-sft}{rinna/bilingual-gpt-neox-4b-instruction-sft}} & MIT & July 2023\\
\quad {\it \href{https://huggingface.co/rinna/bilingual-gpt-neox-4b-instruction-ppo}{rinna/bilingual-gpt-neox-4b-instruction-ppo}} & MIT & August 2023\\
\quad {\it \href{https://huggingface.co/rinna/youri-7b}{rinna/youri-7b}} & LLAMA 2 Community & October 2023\\
\quad {\it \href{https://huggingface.co/rinna/youri-7b-instruction}{rinna/youri-7b-instruction}} & LLAMA 2 Community & October 2023\\
\quad {\it \href{https://huggingface.co/rinna/youri-7b-chat}{rinna/youri-7b-chat}} & LLAMA 2 Community & October 2023\\
\quad {\it \href{https://huggingface.co/rinna/youri-7b-gptq}{rinna/youri-7b-gptq}} & LLAMA 2 Community & October 2023\\
\quad {\it \href{https://huggingface.co/rinna/youri-7b-instruction-gptq}{rinna/youri-7b-instruction-gptq}} & LLAMA 2 Community & October 2023\\
\quad {\it \href{https://huggingface.co/rinna/youri-7b-chat-gptq}{rinna/youri-7b-chat-gptq}} & LLAMA 2 Community & October 2023\\
\quad {\it \href{https://huggingface.co/rinna/nekomata-7b}{rinna/nekomata-7b}} & Tongyi Qianwen & December 2023\\
\quad {\it \href{https://huggingface.co/rinna/nekomata-7b-instruction}{rinna/nekomata-7b-instruction}} & Tongyi Qianwen & December 2023\\
\quad {\it \href{https://huggingface.co/rinna/nekomata-7b-gguf}{rinna/nekomata-7b-gguf}} & Tongyi Qianwen & December 2023\\
\quad {\it \href{https://huggingface.co/rinna/nekomata-7b-instruction-gguf}{rinna/nekomata-7b-instruction-gguf}} & Tongyi Qianwen & December 2023\\
\quad {\it \href{https://huggingface.co/rinna/nekomata-14b}{rinna/nekomata-14b}} & Tongyi Qianwen & December 2023\\
\quad {\it \href{https://huggingface.co/rinna/nekomata-14b-instruction}{rinna/nekomata-14b-instruction}} & Tongyi Qianwen & December 2023\\
\quad {\it \href{https://huggingface.co/rinna/nekomata-14b-gguf}{rinna/nekomata-14b-gguf}} & Tongyi Qianwen & December 2023\\
\quad {\it \href{https://huggingface.co/rinna/nekomata-14b-instruction-gguf}{rinna/nekomata-14b-instruction-gguf}} & Tongyi Qianwen & December 2023\\
\midrule
Language-image model & & & \\
\quad {\it \href{https://huggingface.co/rinna/japanese-clip-vit-b-16}{rinna/japanese-clip-vit-b-16}} & Apache 2.0 & May 2022\\
\quad {\it \href{https://huggingface.co/rinna/japanese-cloob-vit-b-16}{rinna/japanese-cloob-vit-b-16}} & Apache 2.0 & May 2022\\
\quad {\it \href{https://huggingface.co/rinna/japanese-stable-diffusion}{rinna/japanese-stable-diffusion}} & {CreativeML OpenRAIL M} & September 2022\\
\quad {\it \href{https://huggingface.co/rinna/bilingual-gpt-neox-4b-minigpt4}{rinna/bilingual-gpt-neox-4b-minigpt4}} & MIT & July 2023\\
\midrule
Language-speech model & & & \\
\quad {\it \href{https://huggingface.co/rinna/japanese-wav2vec2-base}{rinna/japanese-wav2vec2-base}} & Apache 2.0 & March 2024\\
\quad {\it \href{https://huggingface.co/rinna/japanese-hubert-base}{rinna/japanese-hubert-base}} & Apache 2.0 & April 2023\\
\quad {\it \href{https://huggingface.co/rinna/japanese-hubert-large}{rinna/japanese-hubert-large}} & Apache 2.0 & March 2024\\
\quad {\it \href{https://huggingface.co/rinna/japanese-data2vec-audio-base}{rinna/japanese-data2vec-audio-base}} & Apache 2.0 & March 2024\\
\quad {\it \href{https://huggingface.co/rinna/nue-asr}{rinna/nue-asr}} & Apache 2.0 & December 2023\\
\bottomrule
\end{tabular}
\caption{Based on our vision described in the Introduction~\ref{sec:intro} and Conclusion~\ref{sec:conc}, we are continually releasing Japanese pre-trained models.
This table lists the models we have released by March 2024.}
\label{table:models2}
\end{table*}

\end{document}